\documentclass[conference]{IEEEtran}

\IEEEoverridecommandlockouts

\usepackage{cite}
\usepackage{amsmath,amssymb,amsfonts}
\usepackage{algorithmic}
\usepackage{graphicx}
\usepackage{textcomp}
\usepackage{xcolor}
\def\BibTeX{{\rm B\kern-.05em{\sc i\kern-.025em b}\kern-.08em
    T\kern-.1667em\lower.7ex\hbox{E}\kern-.125emX}}

\usepackage{amsmath}
\usepackage{amssymb}
\usepackage{hyperref}
\usepackage{microtype}
\usepackage{graphicx}
\usepackage{subfig}
\usepackage{booktabs} 
\usepackage{subfig}
\usepackage{multirow}
\usepackage{soul}
\usepackage{amsmath}
\usepackage{amssymb}
\usepackage[utf8]{inputenc}
\usepackage{cleveref}
\newcommand{\floor}[1]{\lfloor #1 \rfloor}

\usepackage{tabularx}
\usepackage{multirow}

\newcommand{\hm}{\textup{HM\,16.20}\,\,}
\newcommand{\vvc}{\textup{VVC}\,}
\newcommand{\ebu}{\textup{EBU\,UHD-1}\,}
\newcommand{\hevc}{\textup{HEVC}\,}

\newcommand{\ie}{\textit{i.e.}}
\newcommand{\eg}{\textit{e.g.}}

\newcommand{\rdoq}{\textup{RDOQ}}
\newcommand{\sq}{\textup{SQ}}
\newcommand{\hmsq}{\textup{HM-SQ}\,\,}

\newcommand{\frdoq}{f_{\textup{RDOQ}}}

\newcommand{\farm}{f_{\textup{ARM}}}
\newcommand{\fcn}{f_{\textup{FCN}}}

\newcommand{\blocksize}{N}

\newcommand{\unquantized}{\boldsymbol{x}}
\newcommand{\refinement}{\boldsymbol{\Delta}}

\newcommand{\quantized}{\boldsymbol{q}}
\newcommand{\quantizedelement}{q}

\newcommand{\scalarquantized}{\quantized_{\sq}}

\newcommand{\quantizedpredicted}{\boldsymbol{\hat{\quantizedelement}}}

\newcommand{\quantizeddelta}{\boldsymbol{\Delta}} 
\newcommand{\quantizeddeltaelement}{\Delta} 

\newcommand{\numtokens}{k}

\def\reviewmode{singleblind}

\IEEEoverridecommandlockouts
\def\BibTeX{{\rm B\kern-.05em{\sc i\kern-.025em b}\kern-.08em
    T\kern-.1667em\lower.7ex\hbox{E}\kern-.125emX}}
    
\IEEEoverridecommandlockouts
\IEEEpubid{\makebox[\columnwidth]{978-1-7281-9320-5/20/\$31.00 © 2020 IEEE  \hfill} \hspace{\columnsep}\makebox[\columnwidth]{ }}

\begin{document}

\title{Parallelized Rate-Distortion Optimized Quantization Using Deep Learning}


\makeatletter
\newcommand{\linebreakand}{%
  \end{@IEEEauthorhalign}
  \hfill\mbox{}\par
  \mbox{}\hfill\begin{@IEEEauthorhalign}
}
\makeatother

\ifnum\pdfstrcmp{\reviewmode}{singleblind}=0{
\author{
    \IEEEauthorblockN{1\textsuperscript{st} Dana Kianfar}
    \IEEEauthorblockA{\textit{Qualcomm AI Research\textsuperscript{*} \thanks{{\textsuperscript{*}Qualcomm AI Research, an initiative of Qualcomm Technologies, Inc. and/or its subsidiaries.}}}\\
        dkianfar@qti.qualcomm.com} 
    \and
    \IEEEauthorblockN{2\textsuperscript{nd} Auke Wiggers} 
    \IEEEauthorblockA{\textit{Qualcomm AI Research\textsuperscript{*}}\\
        auke@qti.qualcomm.com}
    \and
    \IEEEauthorblockN{3\textsuperscript{rd} Amir Said} 
    \IEEEauthorblockA{\textit{Qualcomm AI Research\textsuperscript{*}}\\
        asaid@qti.qualcomm.com}
    \linebreakand 
    \IEEEauthorblockN{4\textsuperscript{th} Reza Pourreza} 
    \IEEEauthorblockA{\textit{Qualcomm AI Research\textsuperscript{*}}\\
        pourreza@qti.qualcomm.com}
    \and 
    \IEEEauthorblockN{5\textsuperscript{th} Taco Cohen} 
    \IEEEauthorblockA{\textit{Qualcomm AI Research\textsuperscript{*}}\\
        tacos@qti.qualcomm.com}
}
}
\else
{
\author{
    \IEEEauthorblockN{1\textsuperscript{st} Anonymous author}
    \IEEEauthorblockA{\textit{Anonymous institution}\\
        Location \\
        Email address} 
    \and
    \IEEEauthorblockN{2\textsuperscript{nd} Anonymous author}
    \IEEEauthorblockA{\textit{Anonymous institution}\\
        Location \\
        Email address} 
    \and
    \IEEEauthorblockN{3\textsuperscript{rd} Anonymous author}
    \IEEEauthorblockA{\textit{Anonymous institution}\\
        Location \\
        Email address} 
    \linebreakand 
    \IEEEauthorblockN{4\textsuperscript{th} Anonymous author}
    \IEEEauthorblockA{\textit{Anonymous institution}\\
        Location \\
        Email address} 
    \and 
    \IEEEauthorblockN{5\textsuperscript{th} Anonymous author}
    \IEEEauthorblockA{\textit{Anonymous institution}\\
        Location \\
        Email address} 
}
}
\fi

\maketitle


\begin{abstract}
    Rate-Distortion Optimized Quantization (RDOQ) has played an important role in the coding performance of recent video compression standards such as H.264/AVC, H.265/HEVC, VP9 and AV1. 
    This scheme yields significant reductions in bit-rate at the expense of relatively small increases in distortion.
    Typically, RDOQ algorithms are prohibitively expensive to implement on real-time hardware encoders due to their sequential  nature and their need to frequently obtain entropy coding costs. 
    This work addresses this limitation using a neural network-based approach, which learns to trade-off rate and distortion during offline supervised training.
    As these networks are based solely on standard arithmetic operations that can be executed on existing neural network hardware, no additional area-on-chip needs to be reserved for dedicated RDOQ circuitry.
    We train two classes of neural networks, a \emph{fully-convolutional network} and an \emph{auto-regressive network}, and evaluate each as a post-quantization step designed to refine cheap quantization schemes such as scalar quantization (SQ). 
    Both network architectures are designed to have a low computational overhead.
    After training they are integrated into the HM 16.20 implementation of HEVC, and their video coding performance is evaluated on a subset of the H.266/VVC SDR common test sequences.
    Comparisons are made to RDOQ and SQ implementations in HM\,16.20.\
    Our method achieves 1.64\% BD-rate savings on luminosity compared to the HM SQ anchor, and on average reaches 45\% of the performance of the iterative HM RDOQ algorithm.
\end{abstract}

\section{Introduction}


A recent development in chip design is the integration of dedicated components for neural network (NN) inference.
With NNs being used in conjunction with or instead of domain-specific algorithms, a neural processing unit (NPU) eliminates the need for domain-specialized hardware that are traditionally present on a system-on-a-chip (SoC). 
For example, in the domain of image processing an end-to-end deep learning approach can achieve state-of-the-art results in low-level denoising and demosaicing tasks, and may outperform a manufacturer's image signal processor (ISP)~\cite{schwartz2018deepisp}.
This is a compelling trend in hardware design as a specialized ISP reserves a considerable amount of area on the SoC.
Additionally, any future efforts directed at optimizing the performance or power consumption of the NPU will not only benefit the \emph{neural} ISP, but also all other processes that use the NPU.

\begin{figure}[t!]
\centering
  \includegraphics[width=\linewidth]{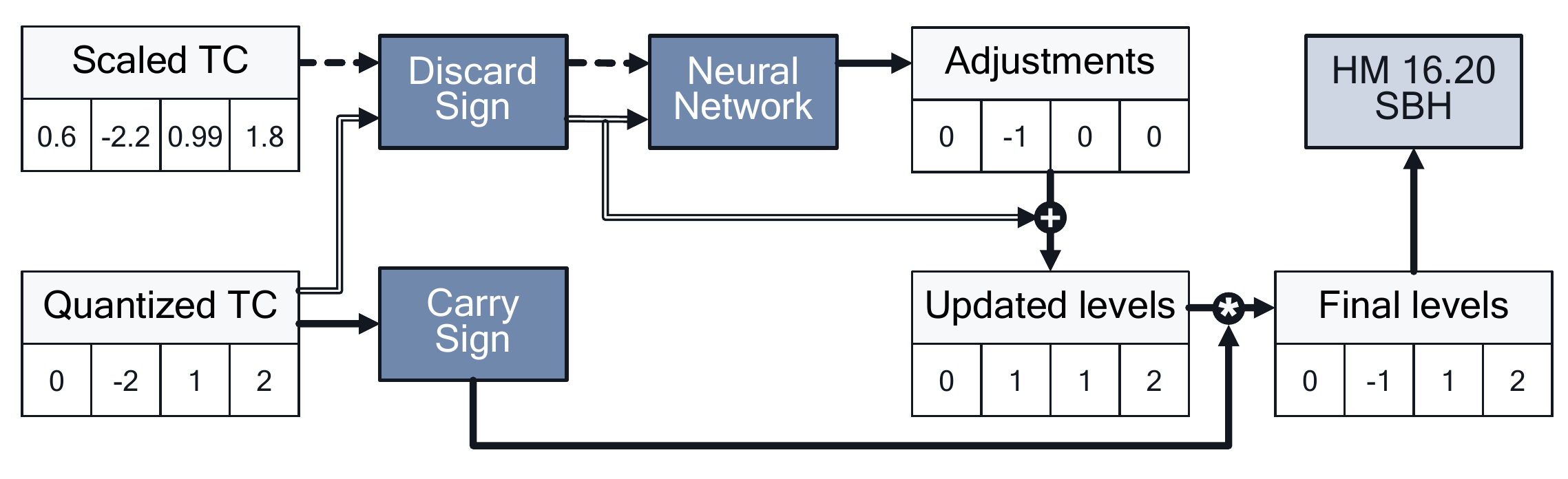}
  \caption{
  Neural network-based Rate Distortion Optimization Quantization. 
  The network takes unsigned scaled transform coefficients (TC) and quantized TCs as input, and predicts an \emph{additive adjustment value} per TC. 
  These adjustments are added to the unsigned quantized TCs, and the sign is re-inserted.
  }
\label{fig:workflow}
\end{figure}

Similarly, the recent video compression standard H.265, commonly referred to as High Efficiency Video Coding (HEVC)~\cite{hevc}, is currently implemented in specialized hardware~\cite{fpga_hevc}.
Recent works have explored replacing or enhancing components of this standard using deep learning, \eg, intra-frame prediction mode decisions~\cite{laude2016deep}, coding tree unit split decisions~\cite{xu2018reducing}, residual prediction~\cite{zhang2017efficient}, and intra-frame rate control~\cite{hu2018reinforcement}. 
In this work, we enhance the \emph{quantization} component of HEVC using deep learning.
The approach is visualized in Fig.~\ref{fig:workflow}.    

\begin{figure*}[ht!]
\centering
  \includegraphics[width=\linewidth]{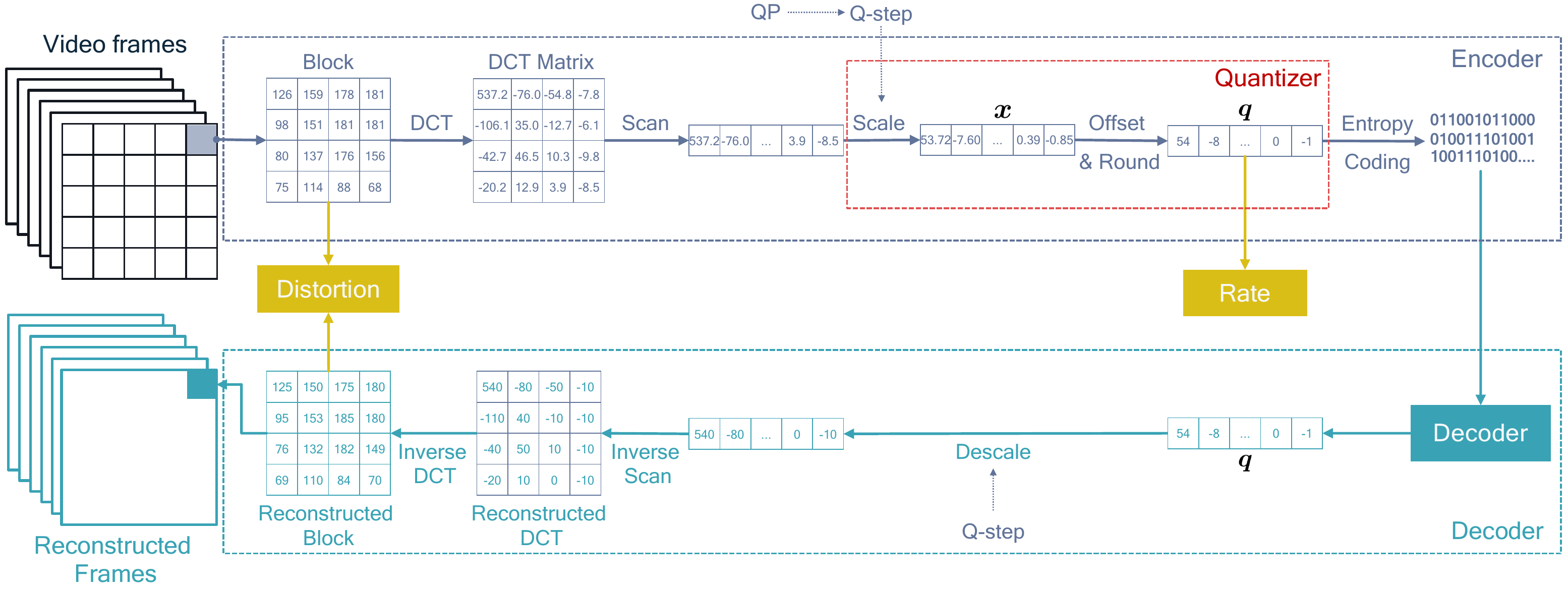}
  \caption{General schematic of the quantizer within the video encoder pipeline in HEVC.}
  \label{fig:codec}
\end{figure*}

HEVC uses a block-based approach where video frames are adaptively partitioned into square coding blocks, as depicted in Fig.~\ref{fig:codec}. 
Compression is achieved by exploiting spatial and temporal redundancies in blocks within and across frames and compressing them using a decorrelating Discrete Cosine Transform (DCT). 
An important step in compression is subsequently performed by during quantization which is the only lossy and irreversible operation in the HEVC pipeline, and thus it is crucial for achieving a good rate-distortion (RD) trade-off.

Rate-distortion optimized quantization (RDOQ) is a standard-compliant quantization procedure that is known to offer a better rate-distortion trade-off than conventional scalar quantization (SQ)~\cite{karczewicz2008rate}. 
RDOQ is a form of adaptive quantization where the quantization scheme is determined by the contents of a block rather than being static.
The method determines optimal \emph{quantization levels} by trading off distortion for bit-rate given some trade-off parameter $\lambda$.
Effectively, RDOQ aims to solve the following discrete optimization problem~\cite{sullivan1998rate} for any given block:
\begin{equation}
\label{eq:rdoq}
    \min_{\quantized} D( \quantized, \boldsymbol{x}) + \lambda \cdot R(\quantized),
\end{equation}
where $\boldsymbol{x}$ is a real vector of scaled transform coefficients (TC) of the block, $\quantized$ is an integer vector of quantization levels with the same dimensionality as $\unquantized$, $D$ is a distortion measure between $\quantized$ and $\boldsymbol{x}$, 
$R(\quantized)$ is the bit-rate of encoding $\quantized$ using a fixed entropy coder, and $\lambda>0$ is a trade-off parameter determined by HM and the user-specified quantization parameter (QP). 

Practical implementations of RDOQ sequentially process blocks of TCs, where for each block many candidate quantization levels are iteratively (\ie, non-exhaustively) optimized for \eqref{eq:rdoq}.
An exact solution to the RDOQ objective in \eqref{eq:rdoq} requires searching over all possible quantization levels for per block of TCs, which is infeasible for real-time video encoding.
However, it is reasonable to anticipate a structure to this optimization problem that can be exploited. 
Previous works have focused on reducing the complexity of RDOQ, for example by determining when a search is unnecessary based on block statistics \cite{zhang2015fast}, by mapping the coefficient distribution to an RD estimate to avoid expensive evaluations during run-time \cite{cui2017hybrid}, or by computing relative differences in RD for pairs of candidates instead of computing their RD separately \cite{highspeedrdoq}. 
Implementations of RDOQ rely on heuristics to keep computational overhead small \cite{karczewicz2008rate}.

These approaches may still be too expensive for real-time encoding due to the iterative and sequential nature of RDOQ as well as any potential frame-rate constraints.
Additionally, a parallelized version may be difficult to implement on hardware and would require reserving additional area on chip for RDOQ circuitry. 
A promising solution direction that has the potential to address both concerns is to train a neural network to imitate an expensive RDOQ algorithm \cite{canh2018rate}.
This approach trades off clock cycles and area-on-chip with multiply-and-accumulate (MAC) operations and energy consumption.
Parallelization is easy to accomplish as most neural network inference hardware is designed to process \emph{batches} of datapoints (\ie, many blocks).

In this work, we train neural networks that determine the quantization level for each TC in a given block, as displayed in Fig.~\ref{fig:workflow}. 
First, we gather high-quality training and validation data using a heuristic search algorithm, applied in succession to RDOQ.
We train and validate two types of neural architectures on this data, namely fully-convolutional neural networks (FCNNs) and autoregressive models (ARMs).
We integrate these networks in HM 16.20, the standard reference implementation of H.265/HEVC, as a quantizer and evaluate their performance on the class-C sequences of the VVC SDR common test conditions \cite{vvc_ctc}. We demonstrate that our models are superior to HM's \emph{scalar quantization with deadzone} baseline by a large margin.
Whereas \cite{canh2018rate} use a FCNN based on an expensive VGG-based semantic segmentation model \cite{long2015fully}, we show that networks with substantially fewer parameters are sufficient for obtaining good performance. 

\section{Methodology}
\label{sec:methodology}

Let $f_{\rdoq}$ be an RDOQ algorithm that maps the block of scaled TCs $\unquantized$ to quantized TCs $\quantized_{\rdoq}$.
Let $\quantized_{\sq}$ be the quantized TCs, also knows as \emph{quantization levels}, obtained by applying \emph{scalar quantization} to $\unquantized$.
Similar to \cite{canh2018rate}, a network is trained to imitate $f_{\rdoq}$ by predicting how each element of $\quantized_{\sq}$ should be adjusted in order to mimic $\quantized_{\rdoq}$. 
In particular, it will be trained to predict discrete adjustment values $\refinement$ such that:
\begin{equation}
    \label{eq:prediction}
    \quantizedpredicted := \refinement + \scalarquantized \approx \quantized_{\rdoq}
\end{equation}
In this way, the network \emph{refines} a content-agnostic quantization $\scalarquantized$ to an improved content-aware quantization $\quantizedpredicted$. 

\subsection{Imitation learning}
\label{sec:methodology:imitationlearning}

We use supervised learning to train a neural network $f$ on data generated using an RDOQ algorithm.
Here, inputs are tuples $(\unquantized, \scalarquantized)$ and the labels are $\Delta$.
Each instance of $\unquantized$ is a block of scaled TCs of size $\blocksize \times \blocksize$, where $\blocksize \in \{4, 8, 16, 32\}$ corresponds to the block dimensions permitted in \hevc.
The last layer of the network will therefore have $\blocksize \times \blocksize \times \numtokens$ outputs, where $\numtokens$ is the number of classes, \ie, the possible adjustment values.
The network output corresponds to the unnormalized log-probability for each adjustment value.

Using stochastic gradient descent, the network $f$ is trained to maximize the probability of selecting the adjustment given by the data, or, equivalently, to minimize the negative log-likelihood $\mathcal{L}$ of the data:
\begin{equation}
    \mathcal{L} = 
        - \mathbf{E}_{\unquantized, \scalarquantized \sim P(\unquantized, \scalarquantized)} \left[ 
            \log P( \quantizeddelta | \unquantized, \scalarquantized ) 
        \right],
\end{equation}
which is the categorical cross-entropy loss as $\quantizeddelta$ is discrete.

Training a network based on examples of expert trajectories is typically referred to as \emph{imitation learning} \cite{schaal1999imitation}.
The performance of the network $f$ is thus bounded by that of the expert (in our case $\frdoq$).

\subsection{Refining training data}
\label{sec:methodology:datacollection}

Training data 
can be obtained by running \hm RDOQ and recording the quantization decisions.
As the performance of the trained network is bounded by the quality of the labels, we propose to spend additional time offline to improve this data using a greedy search method. 
For each $4\times4$ coefficient group in a given block, the procedure exhaustively searches for the optimal quantization levels with respect to \eqref{eq:rdoq}, where the trade-off parameter $\lambda$ is determined by \hm. 
This procedure starts at the coefficient group with the highest variance, located in the top left of the DCT matrix, and proceeds in a raster-scan order. This procedure can be repeated any number of times per block.
The resulting \emph{refined} quantization levels are compared to \hm RDOQ in terms of RD and are saved to the training set only if they are lower.

The search procedure reduces the RD by approximately 0.3\%--1.2\% with respect to HM RDOQ (depending on the quantization parameter of HM), and by 2.6\%--7\% with respect to HM SQ.
Note that while we choose a greedy search procedure, any search algorithm can be used to refine the training data.

\subsection{Network input}
\label{sec:methodology:data_processing}

%

Since quantization decisions are symmetric, only the magnitude of transform coefficients needs to be evaluated.
This simplifies the optimization problem by reducing the number of distinct input-output mappings that the neural network has to learn. 
Therefore, as shown in Fig.~\ref{fig:workflow}, the signs of input variables $(\unquantized, \scalarquantized)$ are discarded before the forward pass of the neural network, and inserted back via multiplication after adding the adjustment values $\refinement$.

Afterwards, all network inputs are standardized by subtracting the empirical mean and dividing by the empirical standard deviation, both obtained from the training set. 
As the input data distributions vary significantly across different quantization parameters (QPs) and prediction structures, we train separate models for each setting.

\subsection{Architectures}
\label{sec:methodology:architectures}

As previously discussed we train feed-forward \emph{fully convolutional} neural networks (FCNNs) and \emph{auto-regressive} models (ARMs).  
Feed-forward models only require a single inference step to obtain predictions for all coefficients in a block.
ARMs, on the other hand, predict adjustment values one-by-one, which allows them to model complex relationships in the data at the expense of a higher computational cost.

\subsubsection{Fully convolutional networks}
\label{sec:methodology:imitationlearning:feedforwardnetwork}

A fully convolutional network $\fcn$, given the scaled TCs $\unquantized$ and quantized TCs $\scalarquantized$, predicts the unnormalized log-probabilities of all $k$ possible adjustment values for each TC simultaneously:
\begin{equation}
    P( \quantizeddelta | \unquantized, \scalarquantized ) \propto \exp( \fcn( \unquantized, \scalarquantized )).
\end{equation}

This approach closely resembles \cite{canh2018rate}. 
Convolutional neural networks are used in many settings and optimized inference on NN-specialized hardware is pervasive.

\subsubsection{Auto-regressive models}
\label{sec:methodology:imitationlearning:autoregressivenetwork}

An ARM is a model that conditions the prediction for adjustment $
\quantizeddeltaelement_i$ on all previous adjustments $\quantizeddelta_{<i}$ given some order (\textit{e.g.}, raster scan), and it estimates a conditional log-probability at each step:
\begin{equation}
    P( \quantizeddeltaelement_{i} | \unquantized, \scalarquantized, \quantizeddelta_{<i} ) \propto \exp ( \farm( \unquantized, \scalarquantized, \quantizeddelta_{<i} )).
\end{equation}

Minimizing the joint negative log-likelihood can be expressed as minimizing the sum of these conditional log-probabilities:
\begin{equation}
    \log P( \quantizeddelta | \unquantized, \scalarquantized ) = 
        \sum_{i=1}^{\blocksize^2} \log P( \quantizeddeltaelement_{i} | \unquantized, \scalarquantized, \quantizeddelta_{<i} ).
\end{equation}
Here, the prediction at step $i=1$ is conditioned only on  $\unquantized$ and $\scalarquantized$ since $\quantizeddelta_{<1} = \emptyset $.

Instead of predicting each $\Delta_i$ in a one-by-one fashion, training of ARMs can be parallelized using \emph{teacher forcing} \cite{bengio2000modeling, larochelle2011neural}, whereby the ground truth $\quantizeddelta_{<i}$ is provided as input during the prediction of $\quantizeddeltaelement_{i}$ for all $i$. 
As a result, the network predicts all adjustment values $\quantizeddelta$ simultaneously. 
However, at test time, when no ground truth is available, the network must be queried once per time-step, which in this case is equal to $\blocksize^2$.
Recent work shows that test time prediction can be accelerated by considering that conditional probabilities may not depend on \emph{all} previous outputs \cite{wiggers2020predictive}.
We utilize this approach here as it has the potential to reduce run-time.

\subsection{Design choices}
\label{sec:methodology:designchoices}

Design choices for training procedure are described below.

\paragraph*{Possible adjustment values} 
The set of adjustment values determine the number of output classes $k$. 
Decreasing the magnitudes of the TCs during quantization (\ie, moving them closer to zero) is often desirable as smaller and more frequent integers can be represented by fewer bits after using an entropy coder.
Therefore, as a heuristic, any set of possible adjustment values should contain \{-1, 0\}.

The empirical distribution of the adjustment values in any training set depends on $\scalarquantized$ and $\quantized_{RDOQ}$.
For our setting, we rarely observe values outside $\{-1, 0, +1 \}$.
Similar to \cite{canh2018rate}, the vast majority of the adjustment values are $0$ which creates a significant class imbalance. 
From a machine learning point-of-view, a larger $k$ makes for a more challenging learning task as the size of the output space increases exponentially and the class imbalance exacerbates.


\paragraph*{Scalar quantization offset} 
A straightforward method to quantize TCs is scalar quantization with offset, whereby a real-valued scalar $c$ is quantized to integer $q$ as follows.
\begin{equation}
    q := \floor{c/s + o}
\end{equation}
Here, $s > 0$ is the quantization step size (Q-step in Fig.~\ref{fig:codec}) and $o \in [0,1]$. This procedure is equivalent to nearest integer rounding (NIR) when $o=1/2$. Note that $x = c/s$ for some DCT coefficient $c$. The Q-step is determined by HM and depends on the user-specified quantization parameter (QP).

NIR is the optimal quantizer for any set of TCs in terms of distortion, but it does not provide a desirable RD trade-off.
The \hm scalar quantization procedure (HM-SQ) uses a smaller offset, which encourages rounding down in order to reduce bit-rate.
We investigate both NIR and HM SQ.

\paragraph*{Loss Scaling} 
Training neural networks to directly optimize RD is not possible as it is impossible to obtain analytical gradients of $R(\quantized)$ with respect to the NN parameters. 
The cross-entropy loss may not be a suitable proxy for RD due to the fact it penalizes all classification errors equally across classes and transform coefficients. 
This symmetry is incompatible with the RD estimators used in video coding as (1) positive and negative changes to the magnitude of a TC may incur different shifts in RD, and (2) DCT transform coefficients corresponds to different basis functions with varying effects on RD. 

In order to alleviate this limitation, we compute a \emph{RD sensitivity map} that scales the cross-entropy loss at each TC according to its empirical effect on RD relative to other TCs. 
Each entry in the map is computed by taking the sum of the absolute value of the change in RD caused by a unilateral shift of a quantized coefficient (\eg, by $-1$ or $+1$), summed over many training samples and finally divided by the mean of said values over the entire map. 

\paragraph*{Zero-masking} 
Due to the nature of the DCT transform and scalar quantization, many quantized TCs are set to zero by HM. 
When a long sequence of quantized TCs are zeros, HM is able to encode the entire sequence efficiently.
As a result, adjusting a quantized TC away from zero could disproportionately increase rate.
A simple heuristic for limiting the effect of erroneous predictions on bitrate is to prevent the network from altering such TCs whose scalar quantized TC is 0. 
It is reasonable to expect that a neural network can learn this heuristic by mapping such inputs to class 0, however this is not naturally enforced by the cross-entropy objective.
\section{Experiments on offline data}

\begin{table}[b!]
    \centering
    \caption{Ablations on QP=22. 
             RD is shown as the percentage reduction with respect to HM-SQ.}
    \label{tab:ablation_input_type}
    \begin{tabular}{l c r r r r}
      \toprule
        &    & \multicolumn{2}{c}{Accuracy (\%)} & \multicolumn{1}{c}{RD (\%)}   \\
         &  Block size & Train  & Validation          & Validation         \\  
      \midrule                                                           
\multirow{3}{*}{\hmsq}           & 4$\times$4   & 99.15 & \bf{98.97} & -3.77      \\
                                 & 8$\times$8   & 99.34 & \bf{99.19} & -0.81      \\
                                 & 16$\times$16 & 99.48 & \bf{99.35} & -0.68      \\
\midrule
\multirow{3}{*}{Nearest integer} & 4$\times$4   & 98.33 & 98.08      & \bf{-4.60} \\
                                 & 8$\times$8   & 98.76 & 98.39      & \bf{-1.96} \\
                                 & 16$\times$16 & 99.69 & 98.62      & -1.91 \\
                                 \midrule
\multirow{3}{*}{Nearest integer + SM}  & 4$\times$4   & 98.34 & 98.07      & -4.59 \\
                                       & 8$\times$8   & 98.71 & 98.37      & -1.94 \\
                                       & 16$\times$16 & 99.01 & 98.63      & \bf{-1.92} \\
      \bottomrule
    \end{tabular}
\end{table}

We evaluate trained networks in terms of RD on a validation set of videos not seen during training. 
Data is collected \emph{offline}, \ie, network decisions are not integrated in HM\,16.20.

\begin{figure}[t!]
\centering
  \includegraphics[width=\linewidth]{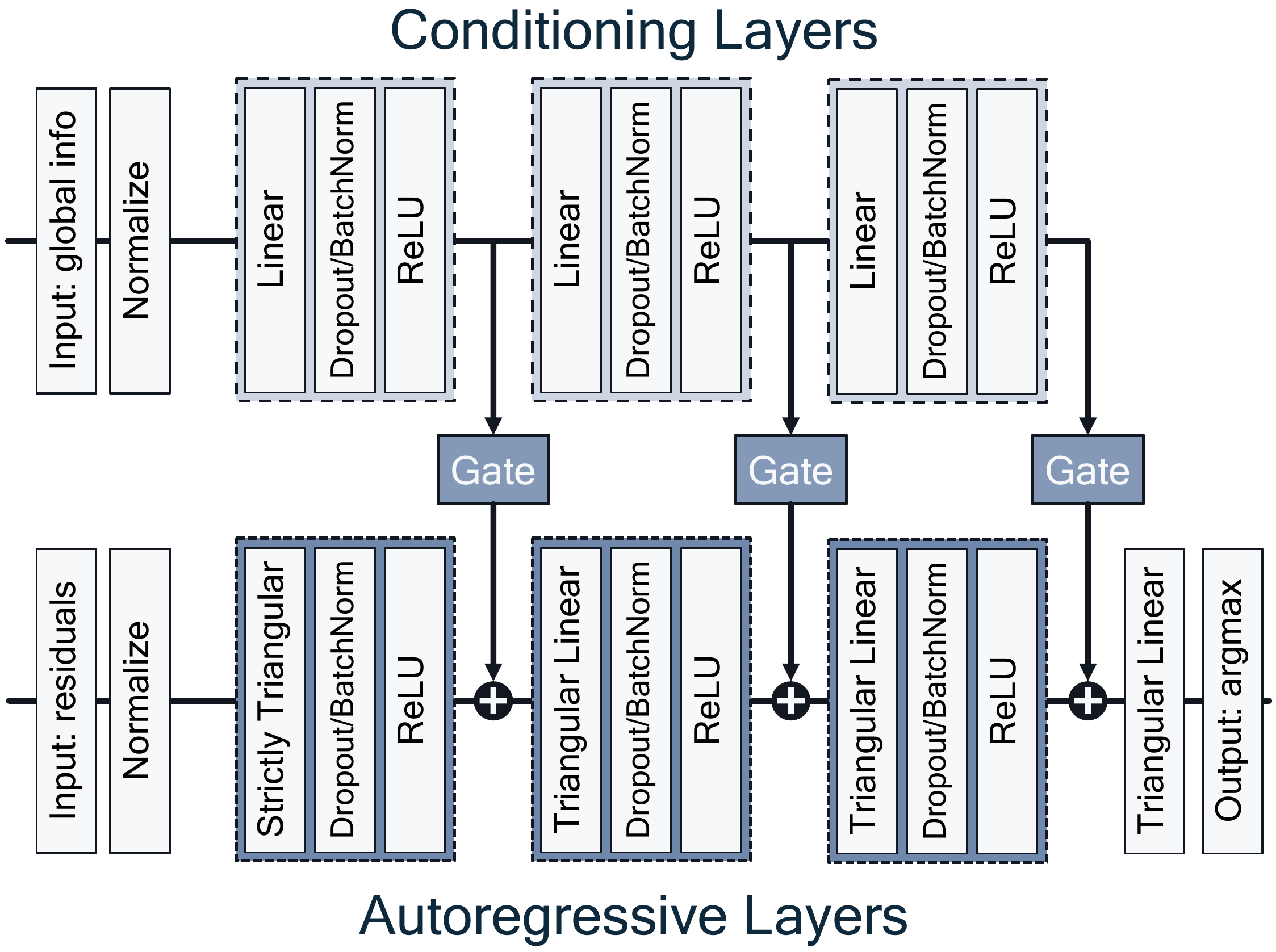}
  \caption{
  The ARM architecture.
  Auto-regressive layers use causal connections via (strictly) triangular layers, ensuring that the temporal dependence is not violated.
  Each auto-regressive layer receives global information (\ie, the quantized and scaled TCs) from the conditioning layers via a fully connected ``Gate'' layer.
  }
  \label{fig:arm-architecture}
\vspace{-3mm}
\end{figure}

\subsection{Dataset}
The \ebu \,test set \cite{ebu_uhd1} is used for training and validation. 
It consists of 12 videos in Ultra-High-Definition consisting of approximately 600-750 frames at 50 frames per second, stored in the 4:2:0 chroma format with 10-bit color depth. 
The videos \textup{StudioDancer} and \textup{ParkDancers} are used for validation and the rest for training. 
To construct the training and validation set, each video is encoded using \hm under different quantization parameters (QP) \{22, 27, 32, 37\}. 
We focus on the all-intra (AI) prediction structure specified under the \textup{intra\_main10} configuration in the \vvc common test conditions \cite{vvc_ctc}. 
We randomly subsample blocks from the encoded videos for training and validation. 
We consider blocks of size {4$\times$4}, {8$\times$8}, and {16$\times$16}.
All-intra prediction was selected to limit the number of experiments, as we train separate networks for different pictures types 
(\eg, i-frames, p-frames).



We limit training and inference to the luma samples (\ie, the Y channel),
motivated by its importance in subjective visual quality, as well as its higher variance and sampling frequency in the encoded bit-stream compared to the UV channels.
By default, we use an offset of $o=\frac{1}{2}$ for the scalar quantization method to obtain network input $ \scalarquantized $ (equivalent to nearest integer rounding).
Early experiments indicated that restricting the adjustment values to $\{-1, 0\}$ yielded the best RD results regardless of the choice of $\scalarquantized$, and we therefore use this configuration throughout. 


\subsection{Network architecture}

\begin{table}[t!]
    \centering
    \caption{Accuracy and RD results on validation data for the FCNN networks. 
             RD is shown as the percentage reduction with respect to HM-SQ.
             }
    \label{tab:validation_results}
    \resizebox{\columnwidth}{!}{%
    \begin{tabular}{c c r r r r r r}
      \toprule
&  &  & \multicolumn{2}{c}{Accuracy (\%)} & \multicolumn{2}{c}{RD (\%)}    \\
      Block size & QP & \# Train data & Train  & Validation            & Validation    & Search     \\  
      \midrule                                                                     
\multirow{4}{*}{4$\times$4}   &  22 & 28.2M &    98.33  &   98.08      &  -4.60    &  -5.59   \\
                              &  27 & 13.1M &    99.05  &   98.91      &  -5.65    &  -6.59  \\
                              &  32 &  7.2M &    99.39  &   99.34	   &  -3.79    &  -4.73   \\
                              &  37 &  3.7M &    99.55  &   99.60	   &  -4.30    &  -5.14  \\
                             \midrule 
\multirow{4}{*}{8$\times$8}   &  22 &  9.6M &    98.76  &   98.39      &  -1.96	   & -3.16     \\
                              &  27 &  5.9M &    99.50  &   99.31	   &  -1.58    & -2.36    \\
                              &  32 &  4.3M &    99.64  &   99.55	   &   -1.72   & -2.51    \\
                              &  37 &  2.7M &    99.75  &   99.73	   &   -2.35   & -3.18   \\
                             \midrule  
\multirow{4}{*}{16$\times$16} &  22 &  1.4M &  99.69    &   98.62      &  -1.91 & -2.86  \\
                              &  27 &  1.1M &  99.02    &   99.46	   &  -1.13 & -1.80 \\      
                              &  32 &  0.9M &  99.73    &   99.68	   &  -1.22 & -1.92  \\
                              &  37 &  0.6M &  99.87    &   99.82	   &  -1.38 & -2.10\\
      \bottomrule
    \end{tabular}
    }

\vspace{-2mm}
\end{table}

The FCNN architecture consists of 2D convolutional layers with a fixed kernel size and channel depth. 
Each layer is followed by batch normalization and a ReLU non-linearity. 
Zero-padding ensures that the output height and width of feature maps remain constant. 
The final layer performs $1\times1$ convolutions and $k$ channels.
Number of hidden layers and channels depend on the block size as follows.
For $4\times4$ FCNNs, we used 3 hidden $3\times3$ Conv2D layers with 256 channels (1.187M parameters), 
for $8\times8$ networks, we used 4 such hidden layers (1.78M parameters), 
and for $16\times16$ network we used 5 hidden layers with 300 channels (3.23M parameters).
In comparison, a standard VGG16 semantic segmentation model has 134M parameters \cite{long2015fully}.

The ARM architecture is shown in Fig.~ \ref{fig:arm-architecture}.
It consists of conditioning layers and auto-regressive layers, which are connected by gated linear layers.
The auto-regressive layers take as input the one-hot encoded adjustments $\quantizeddelta$, whereas the conditioning layers take the same input as the FCNN.
All ARM networks used 3 hidden layers, with 256 hidden units for $4\times4$ networks (0.69M parameters), 384 for $8\times8$ (1.63M parameters) and 512 for $16\times16$ networks (3.42M parameters). 
Batch normalization is used after each hidden layer.
Although we choose the raster scan as the default auto-regressive order, any scan can be used. 
Empirically, we found a negligible difference in performance between ARMs trained on zig-zag scan and ARMs trained on raster scan.

For all experiments, we use the Adam optimizer \cite{kingma2014adam} with learning rate $3 \cdot 10^{-4}$, and default settings $\beta_1 = 0.9$ and $\beta_2 = 0.999$ for the first and second moment terms. An $L2$-norm penalty with a weight of $10^{-6}$ is applied to all parameters.

\subsection{Results}
\label{sec:results}

\begin{table}[t!]
    \centering
    \caption{Accuracy and RD results on validation data for the ARM networks. 
             RD is shown as the percentage reduction with respect to HM-SQ.
             }
    \label{tab:validation_results_arm}
    \resizebox{\columnwidth}{!}{%
    \begin{tabular}{c c r r r r r r}
      \toprule
&  &  & \multicolumn{2}{c}{Accuracy (\%)} & \multicolumn{2}{c}{RD (\%)}    \\
       Block size & QP & \# Train data & Train  & Validation            & Validation    & Search     \\  
      \midrule                                                                     
\multirow{4}{*}{4$\times$4}   &  22 & 28.2M &  98.40 &. 98.05   &  -4.62    &  -5.59   \\
                              &  27 & 13.1M &  99.14    &    98.92   &  -5.72  &  -6.59  \\
                              &  32 &  7.2M &   99.35  &  99.32	 &  -3.79    &  -4.73   \\
                              &  37 &  3.7M &   99.59   &   99.50	   &  -4.29    &  -5.14  \\
                             \midrule 
\multirow{4}{*}{8$\times$8}   &  22 &  9.6M & 98.84    &   98.29 &  -2.01  & -3.16     \\
                              &  27 &  5.9M &  99.42    &   99.20  &  -1.44    & -2.36    \\
                              &  32 &  4.3M &  99.63    &   99.51  &   -1.63   & -2.51  \\
                              &  37 &  2.7M &   99.74   &   99.71  &   -2.28   & -3.18   \\
                             \midrule  
\multirow{4}{*}{16$\times$16} &  22 &  1.4M &  98.71    &     98.08    &  -1.11 & -2.86  \\
                              &  27 &  1.1M &  99.59    &   99.17  &  -0.33 & -1.80 \\      
                              &  32 &  0.9M &   99.77   &   99.47	&  -0.14 & -1.92  \\
                              &  37 &  0.6M &   99.82   &   99.69	   &  -0.03 & -2.10\\
      \bottomrule
    \end{tabular}
    }

\vspace{-2mm}
\end{table}

\paragraph{Ablations}


We first investigate the importance of the input quantization choice on RD. 
Using the FCNN architecture described in Section \ref{sec:results}, we train two sets of networks for the QP=$22$ setting where the quantized TCs are obtained from nearest integer rounding (NIR) and HM-SQ. 
Additionally, we train the same networks with the RD sensitivity map (SM) and display the results in  Table~\ref{tab:ablation_input_type}.

We observe that networks trained on the NIR inputs perform better in terms of RD despite having lower validation accuracy. 
This result is surprising in two ways; (1) HM-SQ by itself is superior to NIR in terms of RD, and (2) networks trained on HM-SQ have higher validation accuracy. 
Both of these discrepancies can be explained by the data imbalance. 
The empirical distribution of $\quantizeddelta$ obtained using NIR is roughly 98\% on class $0$ and 2\% on class $-1$, whereas for HM-SQ it is  99\% and 1\%, meaning that networks can obtain a high accuracy by predicting class $0$ most of the time.
It is evident that the RD SM does not meaningfully change the RD performance.

\paragraph{Main Results}


Results on train and validation data for the FCNN and ARM networks are shown respectively in Tables \ref{tab:validation_results} and \ref{tab:validation_results_arm}. 
Due to the data imbalance which renders the accuracy metric biased, we selected our best performing models using the validation RD scores.
We observe that the RD score is greatest for small blocks, likely due to the fact that these blocks exhibit lower class imbalance than larger blocks because of their smaller output space. 
Additionally, smaller blocks occur more frequently during encoding and therefore provide more training data.
Discrepancies between train and validation accuracy are small implying that no overfitting occurs. 
The corresponding performance of quantization levels found by the heuristic search algorithm is also listed.
We observe that the RD performance of the FCNN and ARM models are nearly tied for $4\times4$ blocks, but the FCNN performs better for other block sizes. 
We conjecture that this discrepancy is due to the difficulty of modelling long sequences using an ARM.

%

Note that the choice of quantizer affects not only the RD but also which blocks are selected by HM.
Therefore, some discrepancy in performance between the offline and \emph{online} setting, where network decisions are integrated in HM, is to be expected.
\section{Integration within HM}

Finally, the models listed in Tables \ref{tab:validation_results} and \ref{tab:validation_results_arm} are integrated in \hm for \emph{online} evaluation.
During encoding, the modified HM implementation queries the network corresponding to the current block size and QP, and selects the adjustment value with the highest predicted probability. 
For $32\times32$ blocks, \hmsq is applied.

We evaluate our model on the class-C sequences of the standard dynamic range (SDR) subset of the VVC test sequences \cite{vvc_ctc}.
For each sequence, we select 10 frames using temporal subsampling with an interleave of 8 frames, using all-intra prediction structure.
In line with \cite{canh2018rate}, network decisions are applied before the \emph{sign bit hiding} (SBH) step of HEVC. 
Note that SBH is also used for both HM-SQ and HM-RDOQ.
As previously mentioned, NN inference is restricted to luma samples (\ie, the Y-channel).

\subsection{Results} 

We present the RD performance expressed as the BD-rate with piece-wise cubic interpolation \cite{BDrate} in Table \ref{tab:main_result_bdrate}. 
The BD-rate uses interpolation to determine the percentage difference in bit-rate under a fixed distortion between a test and reference model. In our case HM SQ acts as reference model.

The network-based RDOQ method outperforms HM SQ for all sequences, reaching 45\% of the on average performance of HM-RDOQ.
Consistent with the pattern we observed on offline holdout data, the FCNN outperforms the ARM.
We do not observe any improvement in BD-rate when using zero-masking (ZM) on the FCNN model.

The main results validate our approach of using deep learning to imitate expensive RDOQ algorithms.


\begin{table}[t!]
    \centering
    \caption{Average BD-rate (\%) for the Y-channel of the VVC SDR class-C sequences with HM-SQ as reference.}
    \label{tab:main_result_bdrate}
    \resizebox{\columnwidth}{!}{%
    \begin{tabular}{r  c c c c}
    \toprule
          Sequence           & HM-RDOQ & ARM & FCNN & FCNN ZM \\ 
     \midrule
          BasketballDrill     & -3.85 &  -0.89 & -1.41 & - \\
          BQMall              & -3.73 &  -1.30 & -1.75 & -1.75 \\
          PartyScene          & -3.30 &  -1.11 & -1.45 & - \\
          RaceHorses          & -3.86 &  -1.41 & -1.97 & -  \\
          \midrule 
          Avg. Class-C        & -3.69 &  -1.18 & -1.64 & - \\
      \bottomrule
    \end{tabular}
    }
    \vspace{-2mm}
\end{table}


%


\section{Conclusion} 

In this work, we demonstrated a neural network-based approach for Rate Distortion Optimized Quantization (RDOQ), a procedure that yields significant reductions in bit-rate at the expense of relatively small increases in distortion.
Two classes of neural networks were investigated and several ablations were performed.
Our networks were trained on the high resolution/frame-rate EBU UHD-1 sequences and were integrated within HM 16.20 and tested on the VVC SDR class-C sequences which have a lower resolution and frame-rate.
Notably, the best-performing networks are of much lower computational complexity than previous works.
Comparisons are made to HM quantization implementations, and both network classes outperform the SQ baseline by a substantial margin. 
 During test time, our networks were able to improve cheap quantizers to perform nearly half as well as HM RDOQ, demonstrating that they are a promising alternative.

\bibliographystyle{IEEEtran}
\bibliography{main}

\end{document}